\pdfoutput=1

\documentclass[11pt]{article}

\usepackage[final]{coling}

\usepackage{times}
\usepackage{latexsym}

\usepackage[T1]{fontenc}

\usepackage[utf8]{inputenc}

\usepackage{microtype}

\usepackage{inconsolata}

\usepackage{graphicx}

\usepackage{graphicx}
\usepackage{boldline} 
\usepackage{booktabs}
\usepackage{multirow}
\usepackage{amsthm} 
\newtheorem{remark}{Remark}
\usepackage{sidecap}
\usepackage{subcaption}
\usepackage{amsmath}
\usepackage{amsfonts}
\usepackage{xcolor}
\usepackage{bm}
\usepackage{hyperref}
\usepackage{rotating}
\usepackage{adjustbox}
\usepackage{algorithm}
\usepackage{algpseudocode} 
\usepackage{array}
\definecolor{green}{rgb}{0.0, 0.5, 0.0}
\definecolor{red}{rgb}{0.82, 0.1, 0.26}
\definecolor{codegray}{gray}{0.9}

\usepackage{pifont}
\newcommand{\cmark}{\ding{51}}%
\newcommand{\xmark}{\ding{55}}%

\newlength{\Oldarrayrulewidth}

\definecolor{darkpurple}{RGB}{75, 0, 130}
\definecolor{darkyellow}{RGB}{230, 184, 0}
\newcommand{\redcomment}[1]{\textcolor{red}{#1}} 
\newcommand{\greencomment}[1]{\textcolor{green}{#1}}

\usepackage[most]{tcolorbox}
\tcbuselibrary{breakable}   

\newtcolorbox{myshadowbox}{
    enhanced,
    colback=white, 
    colframe=black, 
    shadow={1mm}{-1mm}{0mm}{black!50!white}, 
    boxrule=0.5pt 
}
\usepackage{colortbl} 
\usepackage{tabularx} 

\title{A Review of Prominent Paradigms for LLM-Based Agents: Tool Use (Including RAG), Planning, and Feedback Learning}

\author{\textbf{Xinzhe Li} \thanks{Part of work conducted while affiliated with Deakin University.} \\
 Independent Researcher \\
\texttt{sergioli212@outlook.com}
 }

\begin{document}
\maketitle

\begin{abstract}
Tool use, planning, and feedback learning are currently three prominent paradigms for developing Large Language Model (LLM)-based agents across various tasks. Although numerous frameworks have been devised for each paradigm, their intricate workflows and inconsistent taxonomy create challenges in understanding and reviewing the frameworks across different paradigms.
This survey introduces a unified taxonomy to systematically review and discuss these frameworks. 
Specifically, 1) the taxonomy defines environments/tasks, common LLM-profiled roles or LMPRs (policy models, evaluators, and dynamic models), and universally applicable workflows found in prior work, and
2) it enables a comparison of key perspectives on the implementations of LMPRs and workflow designs across different agent paradigms and frameworks. 
3) Finally, we identify three limitations in existing workflow designs and systematically discuss the future work.
Resources have been made publicly available at in our GitHub repository \url{https://github.com/xinzhel/LLM-Agent-Survey}.
\end{abstract}

\begin{figure*}[ht!]
    \centering
    \scriptsize
    \begin{subfigure}[b]{0.26\textwidth}
        \centering
        \includegraphics[width=\textwidth]{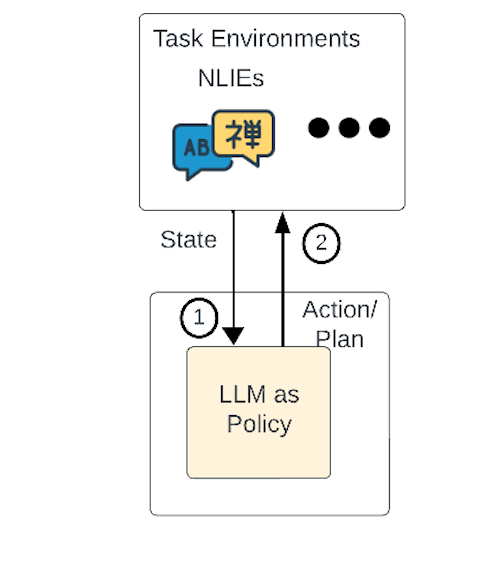}
        \caption{Base Workflows.}
    \end{subfigure} 
    \vspace{0.1cm}
    \begin{subfigure}[b]{0.6\textwidth}
        \centering
        \includegraphics[width=\textwidth]{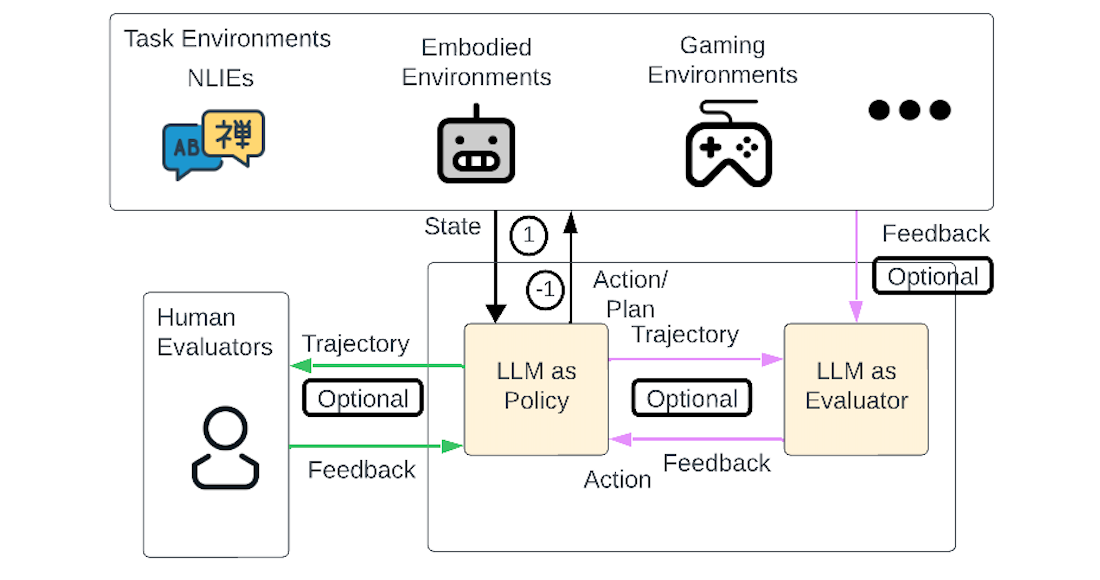}
        \caption{Workflows for Feedback Learning.}
    \end{subfigure} 
    \vspace{0.1cm}
    \begin{subfigure}[b]{0.87\textwidth}
        \centering
        \includegraphics[width=\textwidth]{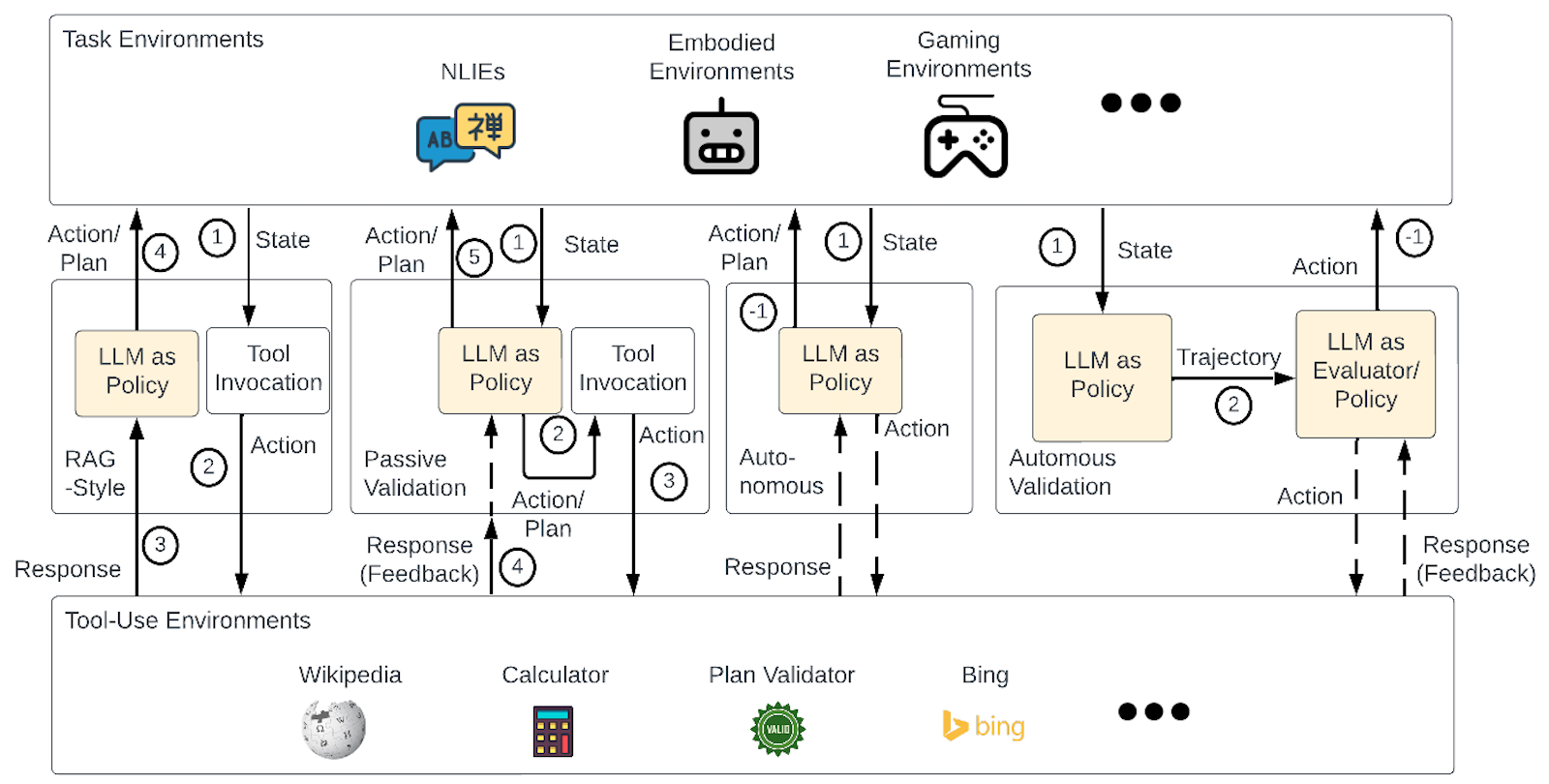}
        \caption{ Workflows for Tool Use, with validation types categorized under both Tool Use and Feedback Learning paradigms.}
    \end{subfigure} 
    \vspace{0.1cm}
    \begin{subfigure}[b]{0.87\textwidth}
        \centering
        \includegraphics[width=\textwidth]{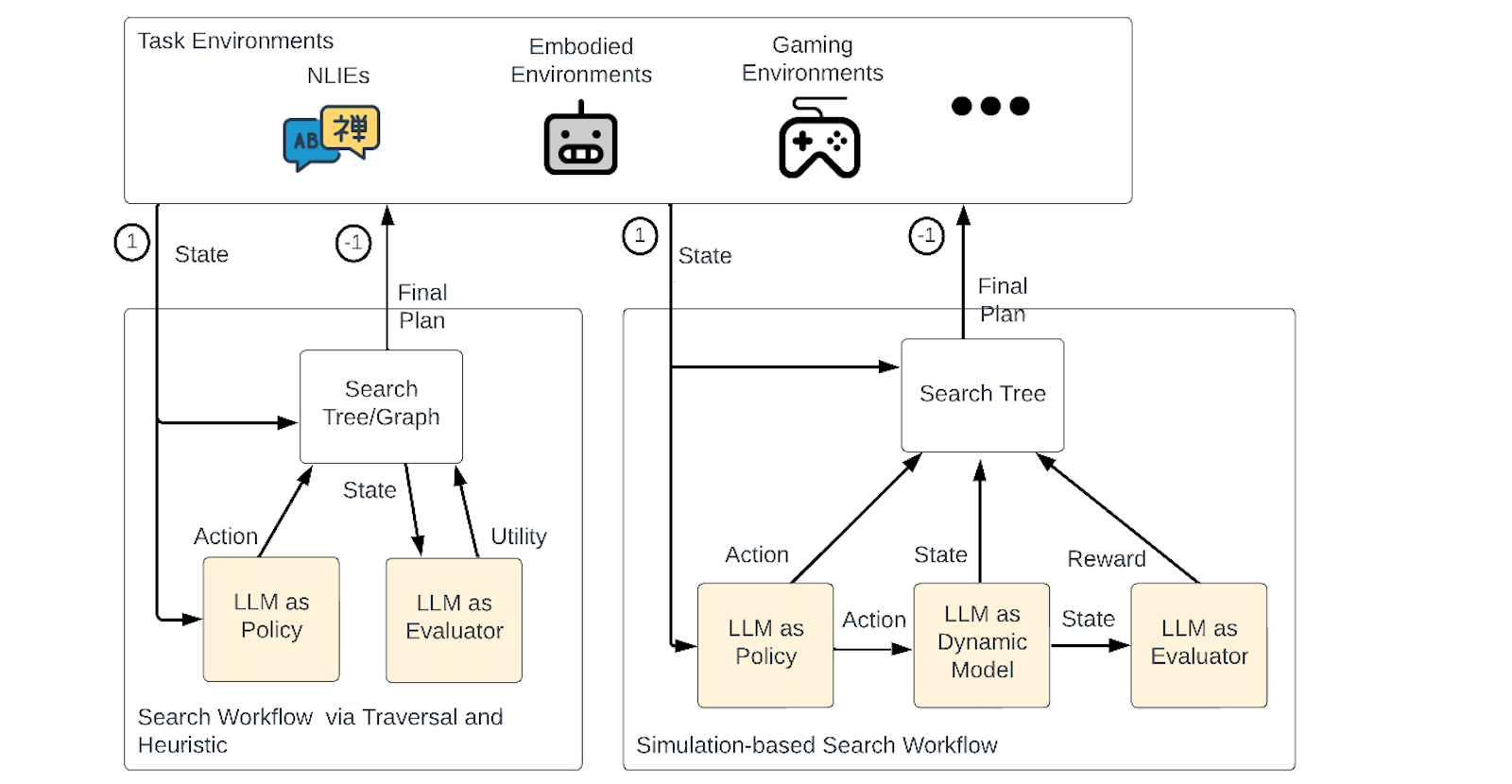}
        \caption{Search Workflows for Planning.}
    \end{subfigure} 
    \caption{Common workflows based on three LLM-Profiled Roles (LMPRs): Policy, Evaluator and Dynamic Model. 
    Numbers in circles indicate the sequence of execution for each step. Unnumbered lines represent iterative steps. 
    Tool-use steps autonomously determined by LMPRs are marked by dashed lines. Circles labeled ``-1'' in the circles indicate the final step. The `Tool Invocation` and `Search Tree/Graph` text boxes indicate that these processes/objects are maintained in agent programs.}
    \label{fig:modular_workflows}
\end{figure*}

\section{Introduction}
Large Language Models (LLMs) have acquired extensive general knowledge and human-like reasoning capabilities \citep{santurkar2023whose,wang2022lsat,zhong-etal-2022-analytical, zhong2023agieval}, positioning them as pivotal in constructing AI agents known as LLM-based agents.
In the context of this survey, LLM-based agents are defined by their ability to interact actively with external tools (\textbf{tool use}) or environments \citep{yao2023react} and are designed to function as integral components of agency for \textbf{planning} \citep{yao2023tree} and \textbf{feedback learning} \citep{shinn2023reflexion}. 

\paragraph{Comparisons with Existing Surveys} 
Current surveys lack a coherent and unified starting point for discussing the three paradigms due to two main limitations:
\textbf{1) Focusing on a specific paradigm or domain}: For example, \citet{huang2024understanding} explore frameworks in the planning paradigm. \citet{hu2024survey,gallotta2024large} examine LLM-based agents in the context of games. In contrast, these paradigms represent general principles and mental models that govern the development and behavior of these agents. Hence, their frameworks should be discussed in a task-agnostic way. To achieve this, we extract universal workflows. Unless otherwise specified, in this survey, the term ``framework'' refers to a complete workflow for specific low-level tasks, while ``workflow'' refers to the task-agnostic process derived from one or more frameworks.
\textbf{2) Lack of a unified basis for comparison}:
Although \citet{wang2024survey} cover all three paradigms, they do not analyze the implementation of algorithmic frameworks in a unified basis.
In contrast, we summarize task-agnostic LLM-profiled roles as the foundation for the development of algorithmic frameworks across different paradigms. Notably,  \citet{wang2024survey} also discuss LLM profiling, but their focus is on personas, which are not relevant to the general roles that underpin these frameworks.
We select four surveys according to their citations on Google Scholar for demonstration purposes, as shown in Table~\ref{tab:survey_comparisons}.
\begin{table*}[h!]
    \centering
    \footnotesize
    \begin{tabular}{p{1.5cm}p{1.25cm}p{1.25cm}|p{1.45cm}p{1.45cm}p{1.25cm}|p{1.25cm}p{1.25cm}|p{1.25cm}}
        \toprule
         & \multicolumn{2}{c}{Tool Use}
         & \multicolumn{3}{c}{Planning}
         & \multicolumn{2}{c}{Feedback Learning}
         & Universal LMPRs
         \\
         & Task-Specific Tools & Workflows 
         & Direct Planning
         & Planning via Search
         & Workflows
         & Feedback Sources
         & Workflows
         \\
        \midrule

        \citet{guo2024large}
        & \xmark & \xmark
        & \xmark & \xmark & \xmark
        & \cmark (\S 3.4) & \xmark
        & \xmark
        \\ \addlinespace
        
        \citet{xi2023rise}
        & \cmark (\S 3.3) & \xmark
        & \cmark (\S 3.1.4) & \xmark & \xmark
        & \cmark (\S 3.1.4) & \xmark
        & \xmark
        \\ \addlinespace

       \citet{huang2024understanding}  
       & \xmark & \xmark
       & \cmark (\S 3) & \cmark (\S 4)  & \xmark 
       & \cmark (\S 6) & \xmark
       & \xmark 
       \\ \addlinespace
       
       \citet{wang2024survey}
       & \cmark (\S 2.1.4) & \xmark
       & \cmark (\S 2.1.3) & \cmark (\S 2.1.3) & \xmark 
       & \cmark (\S 2.1.3) & \xmark 
       & \xmark 
       \\ \addlinespace

       Ours
       & \xmark & \cmark (\S 4.2)
       & \cmark (\S 4.1) & \cmark (\S 4.3) & \cmark
       & \cmark (\S 4.4) & \cmark (\S 4.4) 
       & \cmark (\S 3)
       \\ \bottomrule
    \end{tabular}
    \caption{Coverage of four highly cited surveys and ours. The section numbers of the original manuscripts for the covered perspectives are demonstrated for further reference. We acknowledge the coverage with at least one paragraph focus on that topic.}
\label{tab:survey_comparisons}
\end{table*}
 
\paragraph{Contributions}
This survey offers the following contributions.

\textbf{1) Reviewing the breadth of environments the agents are applicable to (\S \ref{sec:envs})}:
Before investigating LLM-based agents, it is essential to define the scope of universality by summarizing the environments in which these agents operate. 
We categorize the evaluated environments and tasks from the original manuscripts of the reviewed frameworks into two main types: \textbf{decision-making environments} and \textbf{Natural Language Interaction (NLI) environments}, the latter framed from traditional Natural Language Processing (NLP) tasks for agent-based setups.

\textbf{2) Summarizing universal LLM-profiled roles and workflows (\S \ref{sec:lmprs} and \S \ref{sec:workflows})}: 
Workflow designs under planning and feedback-learning paradigms are often based on complex algorithms, such as depth- or breadth-first search \citep{yao2023tree} and Monte Carlo Tree Search \citep{hao-etal-2023-reasoning} for planning, as well as Reinforcement Learning (RL) for feedback learning \citep{shinn2023reflexion}. The complexity increases when these algorithms are adapted to different environments.
We aim to transcend the boundaries of these complex workflows by exploring those that can be universal across all types of environments.
Additionally, we also investigate the workflows for tool use. Such exploration is based on three universal LLM-profile roles.



\textbf{3) Highlighting essential perspectives on LMPRs, workflows, and frameworks (\S \ref{sec:dicussions} and \S \ref{sec:future})}: 
We compare the implementations of LMPRs across various workflow types, emphasize workflows across different paradigms, 
and present alternative viewpoints to our formalization to prevent confusion. Additionally, we critically discuss three limitations of current frameworks and systematically identify potential workflow designs in future research.



\begin{table*}[!ht]
    \centering
    \hyphenpenalty=10000
    \footnotesize
    \begin{tabular}{p{2cm}p{3cm}p{3cm}p{2cm}p{3.7cm}}
    \toprule
    Env Types  &  Entities Interacted With by Agent & Action Properties & Examples of Action Instances & Examples of Env 
    \\ \midrule
     Game \newline Environments 
     & Virtual game elements (objects, avatars, other characters), and possibly other players or game narratives
     & Discrete, 
       Executable,
      \newline Deterministic
     & \redcomment{Move}(\greencomment{Right})
     & 
     BlocksWorld \citep{valmeekam2022large}, 
     CrossWords \citep{yao2023tree}
    \\ \hline
    
    Embodied \newline Environments 
    & Physical world (through sensors and actuators)
    & Discrete, 
      Executable,
      \newline Deterministic
    & \redcomment{Pick\_Up}[\greencomment{Object}] 
    & AlfWorld \citep{shridhar2021alfworld},
    VirtualHome \citep{puig2018virtualhome},
    Minecraft \citep{fan2022minedojo}
    \\ \hline
    
    Web \newline Environments 
     & Virtual web elements
     & Discrete, 
       Executable,
      \newline Deterministic
     & \redcomment{search}(\greencomment{3 ounce bright citrus}), \redcomment{click}(\greencomment{Buy Now})
     & Webshop \citep{yao2022webshop}, WebArena \citep{zhou2024webarena}, AppWorld \citep{trivedi-etal-2024-appworld}
    \\ \hline
    
    NLIEs 
    & Humans (through conversation or text)
    & Free-form,
      Discrete,
      Stochastic 
    &  \redcomment{The answer is} \greencomment{Answer},
    \newline \redcomment{Finish}[\greencomment{Answer}]
    & GSM8K \citet{cobbe2021gsm8k}, HotpotQA \citep{yang-etal-2018-hotpotqa} 
    \\ \bottomrule
    \end{tabular}
    \caption{ Common Task Environments. An action instance is commonly formalized by \redcomment{action predicates} and \greencomment{action arguments}.
    NLIEs refer to Natural Language Interaction Environments. }
    \label{tab:environments_tasks}
\end{table*}

\section{Task Universality}
\label{sec:envs}
This section explores task environments where various workflow types are applicable. 

\subsection{Feedback-Based, Decision-Making Environments}
In this setting, actions yield clear environmental dynamics and rewards for agents to learn from. 
They are typical environments to evaluate planning and RL agents \citep{russell2010artificial,sutton2018reinforcement}.

\paragraph{Rule-Based Game Environments}
These environments, which are deterministic and fully observable, include a variety of abstract strategy games like Chess and Go, and logic puzzles such as the Game of 24 \citep{yao2023tree} and  Blocksworld \citep{hao-etal-2023-reasoning}. They demand deep logical reasoning and strategic planning to navigate and solve.

\paragraph{Embodied Environments}
These settings simulate physical interactions and spatial relationships in the real world. They require agents to engage in navigation, object manipulation, and other complex physical tasks \citep{shridhar2021alfworld,puig2018virtualhome,fan2022minedojo}, reflecting changes in the physical environment. 

\subsection{Information Processing Environments}
\paragraph{Web Environments}
The focus in web-based AI applications is more on processing information and user interaction.
Webshop \citep{yao2022webshop} necessitates a sequence of actions to interact with the environment, such as searching for a product, choosing a color, and clicking ``buy.'' AppWorld \citep{trivedi-etal-2024-appworld} demands a more complex control flow, based on rich programs.

\paragraph{Natural Language Interaction Environments}
Typical NLP tasks are often recontextualized as agentic environments in the study of LLM-based agents \citep{yao2023react,hao-etal-2023-reasoning,yao2023tree}. In this survey, we refer to this setting as Natural Language Interaction Environments (NLIEs) for brevity.
In NLIEs, the environment remains static until the agent acts. 
Unlike typical task environments where natural language serves as an intermediary, in NLIEs, both the states and actions are defined linguistically, making the states conceptual and the actions often ambiguous and broadly defined.
Common setups include: \textbf{1) Single-Step NLIEs}:
Some works \citep{yao2023react, shinn2023reflexion} frame the traditional question-answering (QA) setup as a single-step decision-making process, where the agent generates an answer in response to a question. The process starts with the question as the initial state and concludes when the answer is provided as an action.
Since each trial consists of just one step, there are no environmental dynamics or external feedback. Here, QA encompasses not only typical tasks \citep{cobbe2021gsm8k} but also those often modeled as QA tasks, such as code optimization \citep{shypula2024learning}.
\textbf{2) Deliberate Multi-step NLIEs}:
For tasks without naturally defined intermediate steps, several studies have transformed NLP tasks into a Markov Decision Process to facilitate agentic workflows, e.g., search and planning \citep{yao2023tree,hao-etal-2023-reasoning}.
For example, \citet{hao-etal-2023-reasoning} reformulate subquestions in QA tasks as actions, enabling responses to user queries through a multi-step process. This approach allows the initial question to serve as the beginning of a series of state transitions. Actions may vary from providing direct, free-form answers in single-step QA to strategically proposing subquestions that navigate the agent through sequential updates toward a comprehensive solution. Additionally, \citet{wan2024alphazerolike} suggest that ``splitting an output sequence into tokens might be a good choice'' for defining multi-step NLIEs methodically. Furthermore, \citet{yao2023tree} formulate two-step NLIEs for creative writing by segmenting the problem-solving process into distinct planning and execution phases.

\begin{remark}
In NLP, tasks represent the highest level of abstraction for modeling, whereas in decision-making processes, environments serve this role. It is common to describe an agent as being developed for a specific environment, such as an embodied environment, or for a particular task, such as QA. We refer to it as an NLIE-QA.
\end{remark}

\section{LLM-Profiled Roles (LMPRs)} 
\label{sec:lmprs} 
In this section, we demonstrate three common types of LLM-profiled Roles (LMPRs): policy models, evaluators, and dynamic models. 
They are task-agnostic and commonly used across various workflows.

\paragraph{LLM-Profiled Policy $\text{glm}_{\text{policy}}$}
$\text{glm}_{\text{policy}}$ is designed to generate decisions, which could be an action or a series of actions (plans) for execution in external environments or planning.  
In contrast to typical RL policy models, which learn to maximize cumulative rewards through trial and error, LLM-profiled policy models, denoted as $\text{glm}_{\text{policy}}$, utilize pre-trained knowledge and commonsense derived from extensive textual data. 
We distinguish between two types of $\text{glm}_{\text{policy}}$: an actor $\text{glm}_{\text{actor}}$ directly maps a state to an action, whereas a planner $\text{glm}_{\text{planner}}$ generates a sequence of actions from a given state.

\paragraph{LLM-Profiled Evaluator $\text{glm}_{\text{eval}}$}
$\text{glm}_{\text{eval}}$ provides feedback crucial for different workflows.
During planning, it evaluates each step of actions or the resulting states \citep{hao-etal-2023-reasoning,yao2023tree}, and during feedback learning, they revise the entire decisions \citep{shinn2023reflexion,wang2023describe}. 
Further details are provided in the next subsection.

\paragraph{LLM-Profiled Dynamic Models $\text{glm}_{\text{dynamic}}$}
They predict or describe changes to the environment. Generally, dynamic models form part of a comprehensive world model by predicting the next state $s'$ from the current state $s$ and action $a$. While typical RL uses the probability distribution $p(s' \mid s, a)$ to model potential next states, LLM-based dynamic models directly predict the next state $s' = \text{glm}_{\text{dynamic}}(s, a)$.

\begin{table*}[!ht]
    \footnotesize
    \begin{tabular}{p{1.2cm}p{1.5cm}p{1.5cm}p{1.5cm}p{9cm}} %
    \toprule
          Types & Subtypes
          & Universal LMPRs 
          & Used For
          & Related Frameworks 
          
         \\ \midrule

    Base 
    & $\text{glm}_{\text{actor}}$ 
    & $\text{glm}_{\text{actor}}$
    & /
    &  ReAct \citep{yao2023react}, CoT \citep{wei2022chain}
    \\ \cmidrule{2-5}
    & $\text{glm}_{\text{planner}}$
    & $\text{glm}_{\text{planner}}$
    & Planning
    & \citet{huang2022language}, DEPS \citep{wang2023describe}, Planner-Actor-Reporter \citep{dasgupta2022collaborating}, Plan-and-solve \citep{wang-etal-2023-plan},
    OPEx \citep{shi-etal-2024-opex}
    \\ \hline
    
    Tool-Use 
    & RAG-Style \newline (Passive)
    & $\text{glm}_{\text{policy}}$
    & Tool Use
    & RAG \citep{lewis2020rag,shi-etal-2024-replug,wang2024searching,zhang-etal-2024-multi-task}
    \\ \cmidrule{2-5}
    & Passive Validation
    & Tool Use, Feedback Learning
    & $\text{glm}_{\text{policy}}$
    & \citet{guan2023leveraging}
    \\ \cmidrule{2-5}
    & Autonomous
    & $\text{glm}_{\text{policy}}$
    & Tool-Use
    & MultiTool-CoT \citep{inaba-etal-2023-multitool}, ReAct \citep{yao2023react}, Active RAG \citet{jiang-etal-2023-active} 
    \\ \cmidrule{2-5}
    & Autonomous Validation
    &  $\text{glm}_{\text{policy}}$, $\text{glm}_{\text {eval}}$
    & Tool Use, Feedback Learning
    & CRITIC \citep{gou2024critic}
    \\ \hline
    
    Search 
    & Traversal \newline \& Heuristic
    & $\text{glm}_{\text {policy}}$, $\text{glm}_{\text {eval}}$
    & Planning
    & Tree-of-Thoughts (ToT) \citep{yao2023tree}, Tree-BeamSearch \citep{xie2023selfevaluation}, Boost-of-Thoughts \citep{chen2024boosting},
    Graph-of-Thoughts \citep{besta2023graph}, Tree-of-Traversal \citep{markowitz-etal-2024-tree}
    \\ \cmidrule{2-5} 
    & Simulation-based (MCTS)
    & $\text{glm}_{\text{policy}}$, $\text{glm}_{\text {eval}}$, $\text{glm}_{\text{dynamic}}$
    & Planning
    & RAP \citep{hao-etal-2023-reasoning}, LLM-MCTS \citep{zhao2023large}, \citet{wan2024alphazerolike}, AgentQ \citep{Putta2024AgentQA}, \citet{chen-etal-2024-tree}, \citet{yu-etal-2023-prompt}, \citet{sprueill-etal-2023-monte}
    \\ \hline
    
    Feedback \newline Learning
    & from $\text{glm}_{\text {eval}}$ only
    & $\text{glm}_{\text {policy}}$,
    $\text{glm}_{\text {eval}}$
    & Feedback Learning
    & Reflexion \citep{shinn2023reflexion}, Self-refine \citep{madaan2023self}, TextGrad \citep{yuksekgonul2024textgrad}
    \\ \cmidrule{2-5}
    & from $\text{glm}_{\text {eval}}$ \& Task Env
    & $\text{glm}_{\text {policy}}$,
    $\text{glm}_{\text {eval}}$
    & Feedback Learning
    & Reflexion \citep{shinn2023reflexion}
    
    \\ \cmidrule{2-5}
    & from Humans
    & $\text{glm}_{\text {policy}}$
    & Feedback Learning
    & CRITIC \citep{gou2024critic}
    \\ \bottomrule
    \end{tabular}
    \caption{Universal Workflows of LLM-Based Agents.
    }
    \label{tab:agentic_workflows}
\end{table*}

\section{LMPR-Based Workflows}
\label{sec:workflows}
We explore four types of workflows based on the three types of LMPRs, as illustrated in Figure~\ref{fig:modular_workflows}. 
Table~\ref{tab:agentic_workflows} summarizes the use of these workflows for different paradigms in the prior works.

\subsection{Base Workflows}
The base workflow is as simple as the interaction between $\text{glm}_\text{policy}$ and the environment. These workflows can be categorized based on LLM profiling into two types: planners and actors.
\textbf{1) Planners}:
Many existing frameworks, such as those designed for embodied environments (e.g., \citet{huang2022language}), fall under this category. While some frameworks \citep{dasgupta2022collaborating,wang2023describe} involve complex interactions with task-specific components and low-level, non-LLM-based actors, their universal workflow remains as simple as the base workflow.
For NLIEs, the complete workflow often follows this base model without additional interaction, as seen in \citet{wang-etal-2023-plan}.
\textbf{2) Actors}: Early prompting frameworks for language generation tasks (classified as single-step NLIEs), such as Chain-of-Thought \citep{wei2022chain, kojima2022large}, fit into this category. For embodied tasks, ReAct \citep{yao2023react} employs $\text{glm}_\text{actor}$.

\subsection{Tool-Use Workflows}
We categorize two types of passive workflows and identify two types of autonomous workflows described in previous studies.

\paragraph{RAG-Style Tool Use} A common example of passive tool use is Retrieval-Augmented Generation (RAG) \citep{lewis2020rag}, commonly used in NLIE-QA tasks. In this setup, given a query, a retrieval mechanism collects relevant information to assist $\text{glm}_{\text{policy}}$ in generating a response.

\paragraph{Passive Validation}  \citet{guan2023leveraging} adopt an inverse approach for plan generation. Here, $\text{glm}_{\text{policy}}$ first generates a plan, which is then validated by a separate tool. Depending on the validation outcome, the information may or may not be used to revise the initial plan generated by $\text{glm}_{\text{policy}}$.

\paragraph{Autonomous Tool Use}
In this paradigm, LLMs must be aware of the available tools, which requires including tool information during LLM profiling. The workflow must also handle signals from LMPR generation to invoke tools. Different methods can be applied to enable $\text{glm}_{\text{policy}}$ to autonomously trigger tool usage.
\textbf{1) In-Generation Triggers}: Tools could be invoked during the reasoning process \citep{inaba-etal-2023-multitool, gou2024critic}.
The agent program monitors token generation and pauses when a tool trigger is detected. This pause allows the tool to be invoked, its output processed, and the results integrated into the reasoning process. 
Triggers are defined through tool descriptions, few-shot demonstrations \footnote{See an example prompt in Table \ref{tab:glm_eval_critic}}, or a combination of both \footnote{See an example prompt in Table \ref{tab:glm_actor_multitool}}.
\textbf{2) Reasoning-Acting Strategy}: Introduced by \citet{yao2023react},  each reasoning or acting step completes a full inference cycle, ending with the generation of a stop token. Hence, token-level monitoring is unnecessary. The workflow prompts explicitly for each acting step. 
\textbf{3) Confidence-Based Invocation}: Firstly,  $\text{glm}_\text{policy}$ generates an initial action, and the decision to invoke a tool is based on the confidence level of the generated tokens. \citet{jiang-etal-2023-active} use this method for retrieval invocation, although it is not suitable for general tool use since it cannot specify which tool to invoke. 

\paragraph{Autonomous Validation}
\citet{gou2024critic} utilize $\text{glm}_\text{policy}$ to generate an initial response. The resulting action(s) and the state(s) (i.e., a trajectory) are then passed to  $\text{glm}_\text{evaluator}$, which autonomously determines whether tools should be invoked for validation. 

\begin{remark}
    Tool-use workflows for validation can be viewed as a form of feedback learning, where $\text{glm}_\text{policy}$ receives feedback from the tools. 
\end{remark}

\subsection{Search Workflows}
\paragraph{Traversal and Heuristic-Based Search}
Generations from $\text{glm}_\text{policy}$, instead of direct execution in environments, are used to expand nodes for exploration, stored in a tree or graph structure, such as Tree-Of-Thoughts (ToT) \citep{yao2023tree} and its variants \citep{chen2024boosting, besta2023graph}. $\text{glm}_\text {eval}$ provides a fixed value estimate to select a node for further expansion. 
To expand a tree, ToT applies depth-/breadth-first search (DFS and BFS), while \citet{xie2023selfevaluation} apply beam search.
Notably, the BFS here is functionally equivalent to beam search with $N$ beams as the utility model $\text{glm}_\text{eval}$ is used to maintain the $N$ most promising nodes.
\footnote{Typically, BFS does not rely on a utility model to decide which nodes to expand, since it systematically explores all possible nodes at each level until a terminal state.}

\paragraph{Simulation-Based Search}
Simulation-based search for LLM-based planning agents is often carried out using the classic Monte Carlo Tree Search (MCTS) algorithm \citep{hao-etal-2023-reasoning,zhao2023large,Putta2024AgentQA}. Similar to ToT \citep{yao2023tree}, a tree is built through search and is expanded with $\text{glm}_{\text {policy}}$ and $\text{glm}_\text{eval}$. However, there are two key differences: 
1) Node Selection: The nodes chosen for expansion are determined not only by the static outputs from $\text{glm}_\text{eval}$ or other heuristics (whether they indicate goal attainment), but also by the cumulative statistics accrued over multiple simulations.
Specifically, nodes that lead to better average rewards for subsequent nodes across all simulations (or trajectories) are indeed more likely to be expanded further. 
2) Simulation : Following the selection and expansion phases. A simulation phrase is required where $\text{glm}_{\text{policy}}$, $\text{glm}_{\text{dynamic}}$ and $\text{glm}_{\text{eval}}$ are intimately collaborated, functioning as the roll-out policy. 
Specifically, $\text{glm}_{\text {policy}}$ samples an action $a_t$ given the current state $s_t$, which in turn, is assessed by $\text{glm}_{\text{eval}}$. The top-scoring action is selected, with $\text{glm}_{\text{dynamic}}$ using it to derive $s_{t+1}$, iteratively simulating the trajectory.

\subsection{Feedback-Learning Workflows} 
Within feedback-learning workflows, feedback is fed into $\text{glm}_{\text{policy}}$ for learning. 
One common source of feedback is $\text{glm}_{\text{eval}}$, as in Self-Refine \citep{madaan2023self}.
Other common feedback sources include task environments (e.g., Reflexion \citep{shinn2023reflexion}), tools \citep{gou2024critic,guan2023leveraging} and humans \citet{guan2023leveraging}. Among these, $\text{glm}_{\text{eval}}$ can optionally be used to revise feedback with more contextualized information. In the original manuscript of Reflexion \citep{shinn2023reflexion},
$\text{glm}_{\text{eval}}$ corresponds to ``self-reflection,'' whereas the term ``evaluator'' refers to either heuristics or an LLM-profiled classifier that generates sparse feedback. 
However, this evaluator could be disregarded as a universal LMPR for two reasons:
1) Heuristics are mostly used and lead to better performance, and
2) the evaluator's outputs eventually are fed to ``self-reflection'' for verbal feedback.
When tools are employed to provide feedback \citep{gou2024critic,guan2023leveraging}, the workflow is the same as the tool-use workflow. In this setup, the necessity of invoking tools for feedback is either autonomously determined by $\text{glm}_{\text{eval}}$ \citep{gou2024critic} (See Table~\ref{tab:glm_eval_critic} for an example) or hardcoded \citep{guan2023leveraging}.

\begin{remark}\label{remark:truth_as_fb} 
    \textit{Unrealistic Feedback Source}:
    So far, external feedback from NLIE-QA cannot be realistically generated. Some frameworks use the feedback-learning loop where the feedback source is unrealistic Particularly, \citet{shinn2023reflexion,zhou2024language} use ground truths to generate feedback.
    However, we acknowledge that ground-truth feedback is appropriate when the framework serves as an evaluation baseline, as used by \citet{yao2023tree}.
\end{remark}

\section{Discussions}
\label{sec:dicussions}

\paragraph{Prompting Methods for Profiling}
In the previous subsection, we categorize the original works proposing Chain-of-Thought (CoT) prompting, including zero-shot CoT \citep{wei2022chain} and CoT with few-shot demonstrations \citep{kojima2022large}, under the base workflow since the original work solves task directly via the base workflow. However, these methods themselves can be generalized to different types of LMPRs and workflows, as shown in Appendix~\ref{app:prompting_methods}. 
Some points of the specific use in agents should be highlighted:
1) For planner profiling, zero-shot CoT implementations often fail to produce long-horizon plans \citep{wang2023describe}.
2) To actor profiling under autonomous tool-use workflows, particularly with reasoning-acting strategies, the tool definitions are required to be included in the prompt; and few-shot demonstrations are important to give a clue of generation formats for LLMs including the indication of when to stop.

\paragraph{Workflow Comparisons for Plan Generation}
Both the base workflow using $\text{glm}_\text{planner}$ and search workflows generate a sequence of actions (i.e., a plan). However, they differ fundamentally in how the plans are generated and used.
1) \textbf{Greedy Generation vs. Exploration}: 
The base workflow leverages $\text{glm}_\text{planner}$ to greedily generate a static plan in a single inference step. 
This approach often struggles with long-horizon plans for complex tasks \citep{sun2023adaplanner}.
In contrast, search workflows explore multiple potential solutions and support backtracking, allowing for more robust exploration of options. 
2) \textbf{Plan Execution}: The plans generated by $\text{glm}_\text{planner}$, BFS, DFS, and Beam Search are intended for full execution. However, this can lead to unexecutable actions, especially in the stochastic decision-making environments, if prior actions $(a_1, ..., a_{t-1})$ leads to a state where the next action $a_t$ is invalid (e.g., `standing in front of the fridge` but needing to `open the microwave`). This may cause abrupt interruption in the agent performance. In contrast, in the simulated-absed search via MCTS, only the action at the root node is executed in the actual environment. Subsequent simulated states and actions are discarded, though some implementations may partially retain these states to avoid recomputation. 
This search process repeats after every action taken, continuously recalculating the best action for the updated state.


\paragraph{Workflow Comparisons for Using $\text{glm}_\text{actor}$}
The $\text{glm}_{\text{actor}}$ performs one of three types of actions, depending on the workflow:
1) In base and feedback-learning workflows, $\text{glm}_\text{actor}$ consistently generates task-relevant actions to be executed within the task environment (\textbf{executable task actions}). 
2) In a search workflow, actions generated by $\text{glm}_\text{actor}$ are used for planning algorithms rather than direct execution (\textbf{planning task actions}).
3) In autonomous tool-use workflows, $\text{glm}_{\text{actor}}$ is responsible for generating tool-related actions (\textbf{tool actions}).

\begin{table*}[h!]
    \centering
    \footnotesize
    \begin{tabular}{p{2.4cm}p{3cm}p{3cm}p{6.6cm}}
        \toprule
        Task Formulation 
        & Feedback Types 
        & Applicable Workflows
        & Example Works 
        \\ \midrule
        Text Generation 
        & Free-form reflection 
        & Feedback-learning workflows
        & Self-Refine \citep{madaan2023self}, 
        Reflexion \citep{shinn2023reflexion}, 
        CRITIC \citep{gou2024critic} 
        \\ \hline
        
        Binary/Multi-class
        \newline Classification 
        & Discrete values 
        & Search workflows
        & RAP \citep{hao-etal-2023-reasoning}, 
        Tree-BeamSearch \citep{xie2023selfevaluation},
        ToT \citep{yao2023tree},
        \citet{koh2024tree}
        \\ \hline
        
        Binary Classification 
        & Continuous values (logits)
        & Search workflow via MCTS
        & RAP \citep{hao-etal-2023-reasoning}
        \\ \hline
        
        Multi-choice QA & Choices of top-N actions 
        & Search workflows via traversal and heuristic 
        & ToT \citep{yao2023tree}
        \\ \bottomrule
    \end{tabular}
    \caption{Types of LLM-Profiled Evaluators According to Task Formulation and Feedback Types}
    \label{tab:eval_types}
\end{table*}
\paragraph{Workflow Comparisons for Using $\text{glm}_\text{eval}$} 
In feedback-learning workflows, the generation is passed to $\text{glm}_{\text{policy}}$ for learning, while in search workflows, it is used for planning. 
This fundamental difference (learning vs. planning) leads to the following distinctions: 
\textbf{1) Feedback Use}: As shown in Figure \ref{fig:modular_workflows}, for learning, the output is generated for $\text{glm}_\text{policy}$ to revise and regenerate the entire decision. In search workflows, the output is used to construct a search tree/graph for action selection and further expansion. 
\textbf{2) Feedback Forms}:
These components—$\text{glm}_\text{policy}$ and serch trees/graphs—require different forms of feedback. $\text{glm}_\text{policy}$ processes free-form text, while MCTS utilizes continuous values as rewards, and trees/graphs in BFS, DFS, or beam search rely on discrete values for node selection.
To meet these varying needs, distinct task formulations are employed, as summarized in Table \ref{tab:eval_types}.
\textbf{a) Generating free-form text}: 
$\text{glm}_{\text{eval}}$ is prompted to reflect on previous states and actions, generating reflective text as part of the $\text{glm}_\text{policy}$ prompt in feedback-learning workflows \citep{shinn2023reflexion,gou2024critic}. 
\textbf{b) Binary/multiclass classification}: 
$\text{glm}_{\text{eval}}$ is prompted with specific constraints to generate discrete output tokens, typically ``no'' or ``yes.''
The tokens are converted into scalar values (e.g., 0/1) for use as rewards in MCTS simulations \citep{hao-etal-2023-reasoning}, or to guide decision-making during tree traversal \citep{yao2023tree}.
\textbf{c) Binary classification with scalar values}: This approach differs from the previous one by employing token logits generated by LLMs as scalar feedback values. For example, the probability of a ``yes'' response is computed as:
\[
    v=\text{probs}_{\text{``yes''}} = \frac{e^{l_{\text{``yes''}}}}{e^{l_{\text{``yes''}}} + e^{l_{\text{``no''}}}},
\]
where $l_{\text{``yes''}}$ and $l_{\text{``no''}}$ are the logits for ``yes'' and ``no'' tokens, respectively.
\footnote{Note that such implementations of $\text{glm}_{\text{eval}}$ are less common due to their inaccessibility of state-of-the-art black-box LLMs.} These scalar values can then be used as rewards in MCTS.
\textbf{d) Multi-choice QA}: This formulation is used in scenarios that require selecting from multiple choices, such as choosing from top-N possible actions in traversal-based search workflows \citep{yao2023tree}.



\section{Limitations and Future Works}
\label{sec:future}
\paragraph{Limitation 1: A Unified Solution for Base Workflows and Autonomous Tool-Use Workflows}
ReAct \citep{yao2023react} tries to unify the autonomous tool-use workflow and the base workflow via the reasoning-acting strategy. In this scenario, they implicitly unify tools as a part of task environments, and tool actions and task actions are unified to similar formats. However, the two workflows are still distinct: 1) specifying and comparing different tool-use workflows in a fine-grained manner, 2) distinguishing the two can avoid an illusion that it is universally implementable workflow. 
Specifically, for the latter, the sequence in which reasoning and action outputs alternate is task-dependent.
For QA tasks, the generations of reasoning steps and tool actions are fixed, with alternating prompts for thinking and acting. \footnote{Reasoning-Acting Strategy (QA tasks): See an example prompt in Table \ref{tab:glm_actor_tooluse_hotpotqa}}. 
In contrast, for embodied tasks, the decision whether to proceed with thinking or acting in the next step is autonomously determined by $\text{glm}_{\text{policy}}$ \footnote{Reasoning-Acting Strategy (embodied tasks): See an example prompt in Table \ref{tab:glm_actor_base_alfred}}.

\paragraph{Limitation 2: Lack of Universal Tool-Use Workflow Design}
Another future direction is the development of universal tool use. Although tool use is a universally applicable paradigm, current research tends to focus on using tools for specific tasks such as NLIE-QA or specialized purposes—either enabling $\text{glm}_\text{policy}$ to retrieve new information or enabling $\text{glm}_\text{eval}$ for validation.

\paragraph{Limitation 3: Formulating Language Tasks as Agentic Tasks}
The formulation can help develop a more universally used workflow, e.g., Reflexion \citep{shinn2023reflexion}. However, some deliberate design should be reconsidered for future work, such as the use of ground truths as external feedback in Remark \ref{remark:truth_as_fb}. Another questionable setting is deliberate multi-step NLIEs, e.g., what if a question is simple enough to be answered without the need to be decomposed into subquestions. 

\paragraph{Devising New Workflows}
Our systematic review reveals potential directions for devising new workflows by intertwining existing ones. These workflows can be combined within a single paradigm—for instance, integrating different feedback sources or blending validation-based tool use with autonomous tool use. There is even greater potential for combining workflows across paradigms, such as incorporating optional feedback sources into non-validation tool-use workflows. In fact, validation-style workflows \citep{gou2024critic,jiang-etal-2023-active} inherently merge elements of both the feedback-learning and tool-use paradigms.

\section{Conclusion}
This survey formalizes three common types of LLM-Profiled Roles (LMPRs) and investigates universal workflows for tool use, planning, and feedback learning. Since they are not task-specific, we hope that this will enlighten future research on employing LLM-based workflow designs across different tasks. 
Also, under the shuttle of these LMPRs and workflows, we discuss some nuances of LLM profiling, workflow pros \& cons, and framework implementations.

\section*{Limitations}
Due to the page limit, we can only focus on representative and pioneering work (primarily in * ACL, ICML, ICLR, and NeurIPS) and the perspectives that align with our focus. That means some recent work may be missed, and task-specific components and complete workflow designs are mostly ignored, especially for those frameworks designed for embodied environments. For example, a visual model is normally required to translate pixel observations into textual inputs for $\text{glm}_\text{policy}$ \citep{wang-etal-2023-plan}.
Again, this review aims to summarize the task-agnostic workflows to facilitate a coherent understanding and identify their limitations for future research.

\bibliography{custom}

\appendix
\begin{table*}[h!]
    \centering
    \begin{tabular}{p{12cm}}
    Q: In a dance class of 20 students, $20 \%$ enrolled in contemporary dance, $25 \%$ of the remaining enrolled in jazz dance, and the rest enrolled in hip-hop dance. What percentage of the entire students enrolled in hip-hop dance? \newline
    \redcomment{A: Let's first understand the problem and devise a plan to solve the problem.} \newline
    \redcomment{Then, let's carry out the plan and solve the problem step by step.} \newline
    Plan: 
    \newline
    \begin{myshadowbox} 
    Step 1: Calculate the total number of students who enrolled in contemporary and jazz dance. \newline
    Step 2: Calculate the total number of students who enrolled in hiphop dance. \newline
    Step 3: Calculate the percentage of students who enrolled in hiphop dance. \newline
    \newline
    Solution: \newline
    Step 1: $20 \%$ of 20 students is 4 students. $25 \%$ of the remaining 16 students is 4 students. So, a total of 8 students enrolled in contemporary and jazz dance. \newline
    Step 2: The remaining 12 students enrolled in hip-hop dance. \newline
    Step 3: The percentage of students who enrolled in hip-hop dance is $12 / 20=60 \%$. \newline
    \end{myshadowbox} 
    Therefore, the answer (arabic numerals) is
    \begin{myshadowbox} 
    60\%
    \end{myshadowbox} 
    
    \end{tabular}
    \caption{An LLM planner under NLIE-QA, implemented by zero-shot CoT. The CoT trigger is highlighted in \redcomment{red font}. The content in the shadow box is generated by LLM. Source from \citet{wang-etal-2023-plan}}
    \label{tab:glm_planner_base_gsm8k}
\end{table*}

\begin{table*}[!ht]
    \centering
    \begin{tabular}{p{12cm}}
Interact with a household to solve a task. Here are two examples.
\newline
\{Demo 1\}
\newline
\{Demo 2\}
\newline
Here is the task:
You are in the middle of a room. Looking quickly around you, you see a cabinet 6, a cabinet 5, a cabinet 4, a cabinet 3, a cabinet 2, a cabinet 1, a coffeemachine 1, a countertop 3, a countertop 2, a countertop 1, a drawer 3, a drawer 2, a drawer 1, a fridge 1, a garbagecan 1, a microwave 1, a shelf 3, a shelf 2, a shelf 1, a sinkbasin 1, a stoveburner 4, a stoveburner 3, a stoveburner 2, a stoveburner 1, and a toaster 1. \newline
Your task is to: put a cool tomato in microwave. \newline
>
\begin{myshadowbox}
think: To solve the task, I need to find a tomato, then cool it with the fridge, and finally put it in the microwave.
\end{myshadowbox}
OK.
\newline
>
\begin{myshadowbox}
think: First, I need to find a tomato. A tomato is more likely to be found on the countertop (1-3), cabinet (1-6), drawer (1-3), shelf (1-3), or the sinkbasin. I can start by checking the countertop 1.
\end{myshadowbox}
OK.
> 
\begin{myshadowbox}
go to countertop 1
\end{myshadowbox}
On the countertop 1, you see a dishsponge 1, and a mug 2.
...
    \end{tabular}
    \caption{An LLM actor within base workflow under the Alfred World benchmark, implemented by few-shot CoT. Source from \citep{yao2023react}. }
    \label{tab:glm_actor_base_alfred}
\end{table*}

\begin{table*}[h!]
    \centering
    \begin{tabular}{p{12cm}}
        Use the following tools at your disposal and answer some questions:\\
        - <<Calculator>> \\
        - <<Chemical reaction predictor>> \\
        - <<Molar mass list>> \\
        \\
        Q: Find the amount of $\mathrm{H} 2 \mathrm{CO} 3$ formed on combining 2 moles of $\mathrm{CO} 2$ and 2 moles of $\mathrm{H} 2 \mathrm{O}$.\\
        A: First, use the chemical reaction predictor to determine the chemical equation for the reaction.\\
        Reactants: $2 \mathrm{CO} 2, 2 \mathrm{H} 2 \mathrm{O}$\\
        Products: ?$\mathrm{H}2\mathrm{CO}3$\\
        \redcomment{<<Chemical reaction predictor>>} \\
        $
        2 \mathrm{CO} 2 + 2 \mathrm{H} 2 \mathrm{O} \rightarrow 2 \mathrm{H} 2 \mathrm{CO} 3
        $
        \\
        Second, use the molar mass list to determine the molar mass of $\mathrm{H} 2 \mathrm{CO} 3$. $\mathrm{H} 2 \mathrm{CO} 3$ \redcomment{<<Molar mass list>>} $62 \mathrm{~g} / \mathrm{mol}$\\
        \\
        Third, use the calculator to calculate the amount of 2 moles of $\mathrm{H} 2 \mathrm{CO} 3$. $2 \times 62 =$ \redcomment{<<Calculator>>} $124$\\
        \\
        Therefore, $124 \mathrm{~g}$ of $\mathrm{H} 2 \mathrm{CO} 3$ is formed.\\
         Q: Find the amount of Calcium hydroxide that is required to react with 2 moles of Carbon dioxide to form 2 moles of Calcium carbonate along with 2 moles of Water.\\
         A: 
      
    \end{tabular}
    \caption{An LLM actor within tool-use workflow. In-generation triggers are highlighted in \redcomment{red font}. Source from MultiTool-CoT \citep{inaba-etal-2023-multitool}}
    \label{tab:glm_actor_multitool}
\end{table*}

\begin{table*}[h!]
    \centering
    \begin{tabular}{p{12cm}}
        Solve a question answering task by having a Thought, then Finish with your answer. Thought can reason about the current situation. Finish[answer] returns the answer and finishes the task.\\
        Here are some examples:\\ \\
        \textbf{Question 1:} What is the elevation range for the area that the eastern sector of the Colorado orogeny extends into? \\
        \textit{Thought:} Let's think step by step. The eastern sector of Colorado orogeny extends into the High Plains. High Plains rise in elevation from around 1,800 to 7,000 ft, so the answer is 1,800 to 7,000 ft. \\
        \textbf{Action:} Finish[1,800 to 7,000 ft] \\




        ... \\\\
        \textbf{Question 6:} Were Pavel Urysohn and Leonid Levin known for the same type of work? \\
        \textbf{Thought:} Let's think step by step. Pavel Urysohn is a mathematician. Leonid Levin is a mathematician and computer scientist. So Pavel Urysohn and Leonid Levin have the same type of work. \\
        \textbf{Action:} Finish[Yes] \\
        END OF EXAMPLES)\\ \\ \\ \\
        \textbf{Question}: VIVA Media AG changed it\'s name in 2004. What does their new acronym stand for?
        \\ 
        \textbf{Thought}:
        \begin{myshadowbox}
        Let's think step by step. VIVA Media AG changed its name in 2004. The new acronym must stand for the new name they adopted. To find the answer, we need to research the new name of VIVA Media AG after 2004.Action: Finish[Further research is needed to determine the new acronym.]
        \end{myshadowbox}
        \\
        \textbf{Action}:
        \begin{myshadowbox}
        Finish[Further research is needed to determine the new acronym.]
        \end{myshadowbox}
        
    \end{tabular}
    \caption{An LLM actor within tool-use workflow for the HotPot-QA benchmark, implemented by few-Shot CoT. Source from ReAct \citep{yao2023react}. }
    \label{tab:glm_actor_tooluse_hotpotqa}
\end{table*}

\begin{table*}[h!]
    \centering
    \begin{tabular}{p{12cm}}
        Given a question, please decompose it into sub-questions. For each sub-question, please answer it in a complete sentence, ending with "The answer is". When the original question is answerable, please start the subquestion with "Now we can answer the question:" \\
    \textbf{Question 1:} James writes a 3-page letter to 2 different friends twice a week. How many pages does he write a year? \\
    \textbf{Question 1.1:} How many pages does he write every week? \\
    \textit{Answer 1.1:} James writes a 3-page letter to 2 different friends twice a week, so he writes 3 * 2 * 2 = 12 pages every week. The answer is 12. \\
    \textbf{Question 1.2:} How many weeks are there in a year? \\
    \textit{Answer 1.2:} There are 52 weeks in a year. The answer is 52. \\
    \textbf{Question 1.3:} Now we can answer the question: How many pages does he write a year? \\
    \textit{Answer 1.3:} James writes 12 pages every week, so he writes 12 * 52 = 624 pages a year. The answer is 624. \\
    ... \\ \\

         \textbf{Question 5:} Janet’s ducks lay 16 eggs per day. She eats three for breakfast every morning and bakes muffins for her friends every day with four. She sells the remainder at the farmers' market daily for \$2 per fresh duck egg. How much in dollars does she make every day at the farmers' market? \\

        \textbf{Question 5.1:} \begin{myshadowbox}How many eggs does Janet have left after eating three for breakfast and baking muffins with four?  \end{myshadowbox} \\

      
    \end{tabular}
    \caption{An LLM actor for the GSM8K benchmark. Source from \citet{hao-etal-2023-reasoning}. }
    \label{tab:glm_actor_search_gsm8k}
\end{table*}

\section{Examples Prompts} \label{app:example_prompts}

\subsection{LLM-Profiled Policy Models}
\textbf{1) Base Workflow}:
Prompts for a planner and an actor are demonstrated in Table \ref{tab:glm_planner_base_gsm8k} and \ref{tab:glm_actor_base_alfred}, respectively. The content in the shadow box is generated by LLM. 
\textbf{2) Tool-Use Workflow}: 
Actors under NLIE-QA is demonstrated in Table \ref{tab:glm_actor_multitool} and \ref{tab:glm_actor_tooluse_hotpotqa}. The former shows the implementation of in-generation triggers, while the last one demonstrates the ReAct implementation.
\textbf{3) Search Workflow}:
An actor is required during the expansion stage of MCTS in the RAP workflow. The prompts and expected generations are shown in Table \ref{tab:glm_actor_search_gsm8k}. 


\subsection{Prompts for LLM Evaluators}
\label{app:prompts_for_glm_eval}
Table~\ref{tab:glm_eval_usefulness_gsm8k} and \ref{tab:glm_eval_tot_game24} show LLM evaluators that are profiled as classification tasks,
while Table~\ref{tab:glm_eval_critic} demonstrates another to generate free-form text.

\begin{table*}[h!]
    \centering
    \begin{tabular}{p{12cm}}
        Given a question and some sub-questions, determine whether the last sub-question is useful to answer the question. Output 'Yes' or 'No', and a reason. \\
        \textbf{Question 1:} Four years ago, Kody was only half as old as Mohamed. If Mohamed is currently twice as 30 years old, how old is Kody? \\
        \textbf{Question 1.1:} How old is Mohamed? \\
        \textbf{Question 1.2:} How old was Mohamed four years ago? \\
        \textbf{New question 1.3:} How old was Kody four years ago? \\
        \textit{Is the new question useful? Yes. We need the answer to calculate how old is Kody now.} \\ 
        ... \\\\

        \textbf{Question 5:} Janet’s ducks lay 16 eggs per day. She eats three for breakfast every morning and bakes muffins for her friends every day with four. She sells the remainder at the farmers' market daily for \$2 per fresh duck egg. How much in dollars does she make every day at the farmers' market? \\
        \textbf{New question 5.1:} Now we can answer the question: How much in dollars does she make every day at the farmers' market? \\
        \textit{Is the new question useful? } \\

  
    \end{tabular}
    \caption{An LLM evaluator within simulation-based search workflow for NLIE-QA, implemented by few-shot CoT. It assesses the usefulness of new sub-questions in solving the original question. Source from \citet{hao-etal-2023-reasoning}}
    \label{tab:glm_eval_usefulness_gsm8k}
\end{table*}

\begin{table*}[h!]
    \centering
    \begin{tabular}{p{12cm}}
    Evaluate if given numbers can reach 24 (sure/likely/impossible)
    \newline
    10 14
    \newline
    10 + 14 = 24
    \newline
    sure
    \newline \newline
    ...
    \newline \newline
    1 3 3
    \newline
    1 * 3 * 3 = 9
    \newline
    (1 + 3) * 3 = 12
    \newline
    1 3 3 are all too small
    \newline
    impossible
    \newline \newline
    11, 12
    \end{tabular}
    \caption{An LLM evaluator within Tree-of-Thought Workflow under Game 24, implemented by few-Shot CoT prompting.
    The LLM is profiled for multi-class classification. Source from ToT \citep{yao2023tree}.}
    \label{tab:glm_eval_tot_game24}
\end{table*}

\begin{table*}[h!]
    \centering
    \begin{tabular}{p{12cm}}
    Context: ... \newline
    Sentence: ... \newline
    Is the sentence supported by the context above? \newline
    Answer Yes or No:
    \end{tabular}
    \caption{An LLM evaluator.
    The LLM is profiled for multi-class classification. Source from \citet{manakul2023selfcheckgpt}.}
    \label{tab:glm_eval_zero_shot}
\end{table*}

\begin{table*}[ht!]
    \centering
    \begin{tabular}{p{12cm}}
        \{Few-shot Demonstrations Omitted for Brevity\}\\
        Question: Serianna is a band of what genre that combines elements of heavy metal and hardcore punk? \\
        Proposed Answer: Let's think step by step. Serianna is a band of metalcore genre. Metalcore is a subgenre of heavy metal and hardcore punk. So Serianna is a band of heavy metal and hardcore punk. So the answer is: heavy metal and hardcore punk. \\

        \textbf{1. Plausibility:} 
        \begin{myshadowbox}
        The question asks for the genre that combines elements of heavy metal and hardcore punk, and the answer is "heavy metal and hardcore punk", simply repeat the question. So it's not plausible. \\

        \textbf{2. Truthfulness:} 
        Let's search the question in google: \\
        
        \redcomment{> Search Query:} Serianna is a band of what genre that combines elements of heavy metal and hardcore punk? \\
        > Evidence: \end{myshadowbox}
        \greencomment{[Metalcore - Wikipedia] Metalcore is a fusion music genre that combines elements of extreme metal and hardcore punk.}\\

        \begin{myshadowbox}The evidence suggests that metalcore is a genre that combines elements of extreme metal and hardcore punk, as stated in the proposed answer.\\

        Let's search the proposed answer in google:\\
        \redcomment{> Search Query:} Serianna is a band of metalcore genre. \\
        > Evidence:\end{myshadowbox} 
        \greencomment{[Serianna - Wikipedia] Serianna was a metalcore band from Madison, Wisconsin. The band formed in 2006...}\\

        \begin{myshadowbox}The evidence suggests Serianna is a metalcore band, the proposed answer is correct about this.\\

        Above all, the proposed answer correctly identifies that Serianna is a band of the metalcore genre, which combines elements of heavy metal and hardcore punk. However, the final answer is not plausible since it just lists the genres that metalcore combines.\end{myshadowbox}
    \end{tabular}
    \caption{An LLM evaluator within the Feedback-Learning workflow (feedback from tools). In-generation triggers are highlighted in \redcomment{red font}, and tool-generated content is highlighted in \greencomment{green font}. Source from \citet{gou2024critic}.}
    \label{tab:glm_eval_critic}
\end{table*}

\subsection{Prompts as Dynamic Models}
Table~\ref{tab:glm_dynamic} is profiled as a dynamic model.
\begin{table*}[h!]
    \centering
    \begin{tabular}{p{12cm}}
        Given a question, please decompose it into sub-questions. For each sub-question, please answer it in a complete sentence, ending with "The answer is". When the original question is answerable, please start the subquestion with "Now we can answer the question: ".
        \\\\
        \textbf{Question 1:} Weng earns \$12 an hour for babysitting. Yesterday, she just did 50 minutes of babysitting. How much did she earn? \\
        \textbf{Question 1.1:} How much does Weng earn per minute? \\
        \textbf{Answer 1.1:} Since Weng earns \$12 an hour for babysitting, she earns \$12 / 60 = \$0.2 per minute. The answer is 0.2. \\
        \textbf{Question 1.2:} Now we can answer the question: How much did she earn? \\
        \textbf{Answer 1.2:} Working 50 minutes, she earned \$0.2 x 50 = \$10. The answer is 10. \\\\
        ...
        \\\\
        \textbf{Question 5:} Janet’s ducks lay 16 eggs per day. She eats three for breakfast every morning and bakes muffins for her friends every day with four. She sells the remainder at the farmers' market daily for \$2 per fresh duck egg. How much in dollars does she make every day at the farmers' market? \\
        \textbf{Question 5.1:} How many eggs does Janet have left after eating three for breakfast and using four for muffins? \\
        \textbf{Answer 5.1:}  \\
    \end{tabular}
    \caption{An LLM-Profiled Dynamic Model.}
    \label{tab:glm_dynamic}
\end{table*}

\section{Prompting Methods of LLM-Profiled Roles}
\label{app:prompting_methods}
The use of different prompting methods are summarized in Table~\ref{tab:prompting_methods} .
\begin{table*}[!h] 
\centering 
\footnotesize
\begin{tabular}{p{1cm}p{1.5cm}p{10cm}p{2.5cm}}
    \toprule
     & Prompting  & Example Works & Example Prompts (in Appendix)  
    \\ \midrule
    $\text{glm}_{\text{actor}}$
    & Few-shot 
    & ReAct \citep{yao2023react}, 
    Reflexion \citep{shinn2023reflexion}, 
    RAP \citep{hao-etal-2023-reasoning},
    MultiTool-CoT \citep{inaba-etal-2023-multitool} 
    & Table \ref{tab:glm_actor_multitool}, \ref{tab:glm_actor_tooluse_hotpotqa}
    \\ \hline
    
    $\text{glm}_{\text{planner}}$ 
    & Zero-shot 
    & Plan-and-Solve \citep{wang-etal-2023-plan}, 
    LLM Planner \citep{huang2022language} 
    & Table \ref{tab:glm_planner_base_gsm8k}
    \\ \cmidrule{2-4}

    & Few-shot 
    & DEPS \citep{wang2023describe}, Planner-Actor-Reporter \citep{dasgupta2022collaborating} 
    \\ \hline
    
    $\text{glm}_{\text{evaluator}}$ 
    & Few-shot
    & RAP \citep{hao-etal-2023-reasoning}, 
    Tree-BeamSearch \citep{xie2023selfevaluation}, 
    Reflexion \citep{shinn2023reflexion},
    CRITIC \citep{gou2024critic}
    & Table~\ref{tab:glm_eval_usefulness_gsm8k}, \ref{tab:glm_eval_tot_game24}
    \\ \hline
    $\text{glm}_{\text{dynamic}}$ 
    & Few-shot
    & RAP \citep{hao-etal-2023-reasoning} 
    & Table~\ref{tab:glm_dynamic} 
    \\ \bottomrule
\end{tabular}
\caption{Prompting Methods of LLM-Profiled Roles} 
\label{tab:prompting_methods}
\end{table*}


\end{document}